\documentclass{article}
\usepackage[preprint]{neurips_2023}

\usepackage[utf8]{inputenc}
\usepackage[T1]{fontenc}
\usepackage{hyperref}
\usepackage{url}
\usepackage{booktabs}
\usepackage{amsfonts,amsmath,amssymb}
\usepackage{mathtools}
\usepackage{nicefrac}
\usepackage{microtype}
\usepackage{graphicx}
\usepackage{xcolor}
\usepackage{enumitem}
\usepackage{algorithm}
\usepackage{algpseudocode}
\graphicspath{{figures/}}
\newcommand{\inputif}[2]{\IfFileExists{#1}{\input{#1}}{\emph{#2}}}
\newcommand{\figfile}[4]{\IfFileExists{#1}{%
  \begin{figure}[t]\centering\includegraphics[width=#2\linewidth]{#1}%
  \caption{#3}\label{#4}\end{figure}}{}}
\newcommand{\figfileh}[4]{\IfFileExists{#1}{%
  \begin{figure}[h]\centering\includegraphics[width=#2\linewidth]{#1}%
  \caption{#3}\label{#4}\end{figure}}{}}
  
\title{An Agentic AI Scientific Community for\\ Automated Neural Operator Discovery}

\author{%
  Luis Loo and Ulisses Braga-Neto \\[1ex]
  Department of Electrical \& Computer Engineering\\
  Texas A\&M University, College Station, TX\\
  \texttt{loo,ulisses@tamu.edu}
}

\begin{document}
\maketitle

\begin{abstract}
We present an agentic approach to autonomous neural operator discovery based on an \emph{AI scientific community}, which  consists of a swarm of virtual laboratories that interact under a citation-based economy of influence. Highly-cited labs found new labs that follow their research direction and replace non-performing labs. Each virtual lab contains three agents: an LLM planner that proposes an architecture, a numerical worker that trains and measures it, and an LLM reviewer that participates in cross-lab peer review. All labs share a common vocabulary consisting of DeepONet (branch-trunk), Fourier, Transformer (attention), wavelet, and residual convolutional neural operator building blocks. We evaluate the neural operator AI scientific community on five problems, namely piecewise regression, the linear advection and Burgers 1D PDEs, and the Navier-Stokes and Darcy flow 2D PDEs, while repeating the simulation three times for each problem. The results show that the neural operator AI scientific community is capable of discovering high-accuracy, low-parameter-count neural operator architectures. All 9{,}623 LLM calls are logged and audited, which reveals that the virtual lab LLM planners choose to hybridize in 99.8\% of their logged decisions, consistently returning multi-family hybrids. Moreover, we conducted an ablation study by replacing the LLM agents in each lab by rule-based alternatives, which caused
the scientific community to collapse to non-hybridized single-family stacks in several cases, showing that LLM agency is needed to preserve diversity. The results suggest a no-free-lunch theorem for neural operators: \emph{there is no universal winner}. The code, configurations, and the complete LLM transcripts are released at \url{https://github.com/luislootx/AI-SC}.
\end{abstract}

\section{Introduction}
Operator learning has become a central tool for building fast surrogates of partial differential equation (PDE) solvers~\cite{kovachki2023neuraloperator}. However, each operator family encodes a fixed inductive bias: global spectral convolutions for the Fourier Neural Operator (FNO)~\cite{li2021fno}, a branch-trunk decomposition for DeepONet~\cite{lu2021deeponet}, multiresolution bases for wavelet operators~\cite{tripura2023wno}, and attention for transformer-style operators~\cite{cao2021galerkin,hao2023gnot}. In addition, selecting the architecture hyperparameters, such as depth, width, and block composition, for a new PDE is still done by hand.
This raises a natural question:

\begin{quote}
\emph{Can a simulation of a neural operator research community consisting of automated agentic virtual labs discover the right neural operator architecture for a given problem with no human prior or intervention?}
\end{quote}

\figfile{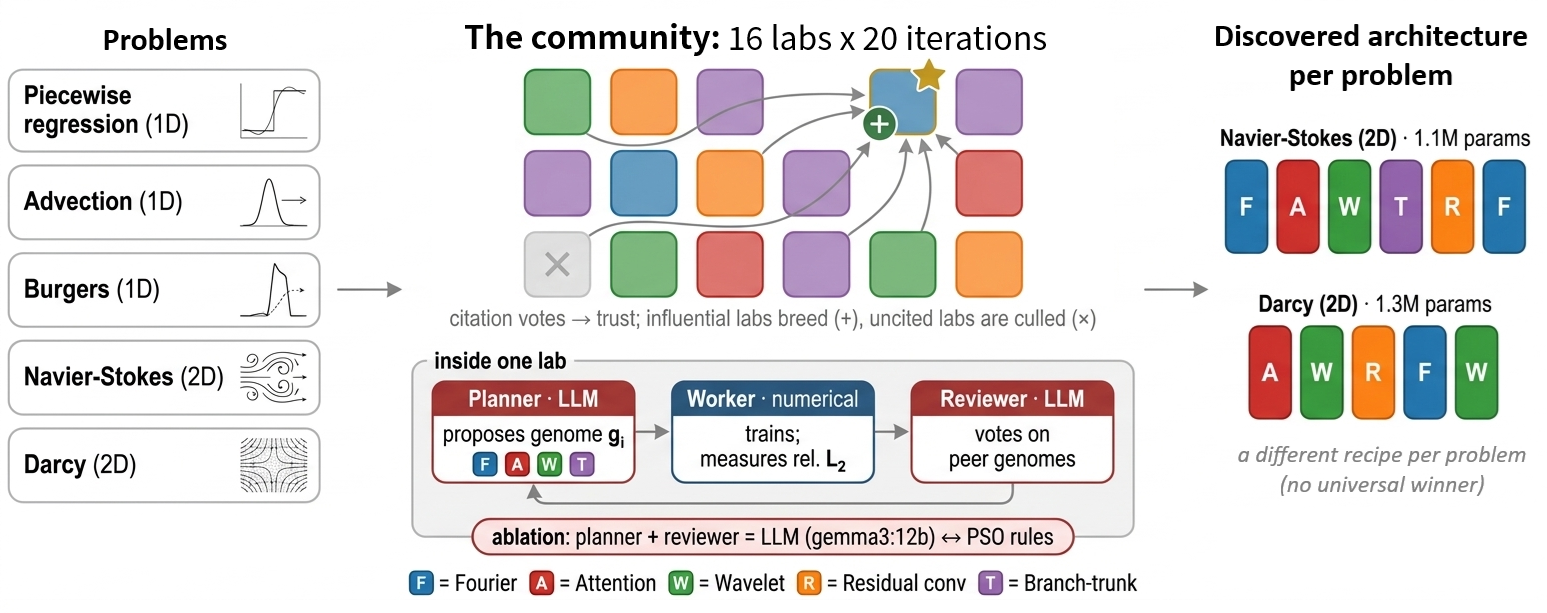}{1.0}
{The neural operator AI scientific community.}
{fig:method}

To answer this question, we employ an \emph{AI scientific community} framework, which combines agentic AI with swarm intelligence to create decentralized networks of virtual laboratories~\cite{braganeto2026aicommunity}. In this paradigm, each particle in the swarm represents a complete virtual laboratory instance, enabling collective scientific exploration that mirrors real-world research communities. This framework\linebreak leverages the inherent properties of swarm intelligence, namely, decentralized coordination,\linebreak balanced exploration-exploitation trade-offs, and emergent collective behavior. The swarm of virtual laboratories explores the design space under a citation-based economy, where well-cited labs gain influence and breeding priority while uncited labs are replaced.

\paragraph{Main Contributions.}
\begin{enumerate}[leftmargin=1.5em,itemsep=2pt]
\item \textbf{An agentic framework for neural operator discovery.} We present an AI scientific community for neural operator architecture search, with LLM planner and
reviewer agents and a gradient-based numerical worker.
\item \textbf{An ablation study of LLM agency.} We replace the LLM planners and reviewers in each lab by rule-based agents, isolating
what the language model adds under an identical training budget. The answer is
architectural rather than accuracy, where the research community with LLM planners propose parameter-lean multi-family
hybrids, whereas the community with rule-based agents tends to collapse around canonical families. 
\item \textbf{A no-free-lunch result for neural operators.} The results of the simulation show that no operator is universal. The AI scientific community with LLM agents discovers multi-family hybrid architectures, whereas the ablated community with rule-based agents tends to collapse to canonical single-family stacks, Fourier on the spectral-friendly Burgers and Navier–-Stokes problems and wavelet on the piecewise regression and Darcy problems. 
\end{enumerate}

\section{Related Work}

\paragraph{Neural operators.} FNO~\cite{li2021fno} learns a convolution kernel in Fourier
space, whereas DeepONet~\cite{lu2021deeponet} separates input-function encoding (branch) from query-location encoding (trunk). Rigorous comparisons find FNO and DeepONet each win in different regimes~\cite{lulu2022comparison}, suggesting that no specific neural operator architecture dominates uniformly over all problems, a fact that the neural operator AI scientific community rediscovers in a much broader context, which includes  wavelet operators with multiresolution bases~\cite{tripura2023wno}, transformer/attention operators~\cite{cao2021galerkin,hao2023gnot}, and convolutional neural operators with residual blocks~\cite{raonic2023convolutional}. Other variants include U-shaped multiscale operators~\cite{rahman2023uno} and geometry-adapted variants for irregular domains~\cite{li2023geofno}.

\paragraph{Architecture search.} Neural architecture search (NAS) automates model
design via reinforcement learning~\cite{zoph2017nasrl}, evolutionary search~\cite{real2019regevo}, gradient-based relaxation of the search space~\cite{liu2019darts}, and many other strategies~\cite{elsken2019nassurvey}. A more radical line evolves entire learning algorithms from primitive operations rather than selecting among predefined blocks~\cite{real2020automlzero}. The neural operator AI scientific community performs an evolutionary/particle-swarm  search~\cite{kennedy1995pso} over an operator-block genome, where the proposal and selection policies are delegated to LLM agents.

\paragraph{LLMs as optimizers and architecture proposers.} Large language models have
been used as black-box optimizers~\cite{yang2024opro}, as proposal engines for neural architecture search~\cite{zheng2023genius}, and for program- and hypothesis-search in mathematical and scientific discovery~\cite{romeraparedes2024funsearch}. Closest to our LLM planner is EvoPrompting~\cite{chen2023evoprompting}, which uses a language model as the mutation and crossover operator inside an evolutionary loop to propose code-level architectures. Our planner plays the same role over an operator-block genome, but we make two design choices that sharpen measurability: the fitness signal comes from a deterministic training procedure rather than a learned predictor, and every LLM-driven run is paired with a rule-based arm so the language model's contribution is isolable by ablation.

\paragraph{Multi-agent LLM systems.} A growing body of work proposes {\em virtual labs}, which are teams of LLM agents with distinct roles that communicate to solve a task, e.g., conversational
multi-agent frameworks~\cite{wu2024autogen}, communicative agent societies~\cite{li2023camel}, and role-specialized teams that emulate a software company~\cite{qian2024chatdev} or a standardized-operating-procedure organization~\cite{hong2024metagpt}. At the workflow level such teams drive end-to-end research agents~\cite{lu2026aiscientist} and numerical algorithm design~\cite{toscano2025athena}. Our framework is different since it takes this paradigm one level up by employing a swarm of virtual labs. Each lab consists of planner, worker, and reviewer agents. The reviewer's accept/reject vote is a form of LLM-as-judge evaluation~\cite{zheng2023llmjudge}, here applied to peer review of candidate operators rather than to chatbot responses. The AI scientific community itself was proposed in \cite{braganeto2026aicommunity}.

\section{Problem Formulation}\label{sec:problem}
\paragraph{Operator learning.} Given a PDE with solution operator
$\mathcal{G}:\mathcal{A}\to\mathcal{U}$ mapping an input function $a\in\mathcal{A}$ (e.g.\ an initial condition or coefficient field) to the solution $u=\mathcal{G}(a)\in\mathcal{U}$, we seek a parametric surrogate $\mathcal{G}_\theta$ minimizing the expected relative $L_2$ error
\begin{equation}
\mathcal{L}(\theta) \;=\; \mathbb{E}_{a\sim\mu}
\frac{\lVert \mathcal{G}_\theta(a) - \mathcal{G}(a)\rVert_2}
     {\lVert \mathcal{G}(a)\rVert_2},
\end{equation}
estimated on sampled input-output pairs. The relative $L_2$ error is our primary accuracy metric throughout.

\paragraph{Architecture search space.} We define an architecture by a
\emph{genome} $g=(b_{1:L},\,c,\,m,\,\sigma,\,\dots)$, where $b_{1:L}$ is a sequence of $L\in[2,8]$ blocks drawn from a vocabulary
\begin{equation}
\mathcal{B}=\{\textsc{branch\_trunk},\textsc{Fourier},\textsc{attention},\textsc{wavelet},
\textsc{residual\_conv}\},
\end{equation}
$c$ is the hidden width, $m$ the number of spectral modes, $\sigma$ the activation, plus gating, skip connections, dropout, and learning rate. A genome compiles to a concrete neural operator $\mathcal{G}_{\theta(g)}$. The discovery problem is the bilevel program
\begin{equation}
g^\star \;=\; \arg\min_{g}\; \mathcal{L}\big(\theta^\star(g)\big),
\qquad
\theta^\star(g) \;=\; \arg\min_{\theta}\; \widehat{\mathcal{L}}_{\text{train}}\big(\theta; g\big),
\end{equation}
where the inner problem is ordinary gradient-based training of a fixed architecture and the outer problem searches genomes. We solve the outer problem with the swarm of agentic virtual labs, while the inner problem is solved by gradient descent.

\section{Methodology}\label{sec:method}

\paragraph{AI scientific community framework.}\label{sec:framework}
A swarm of $N$ virtual labs explores the genome space. Each lab $i$ holds a current genome $g_i$, a personal best, a velocity (for the analytic update), and a \emph{trust} score accumulated through peer review. Each iteration consists of four phases (see Algorithm~\ref{alg:swarm}):

\begin{enumerate}[leftmargin=1.5em,itemsep=1pt]
\item \textit{Planning.} Each lab proposes its next genome $g_i$.
\item \textit{Training.} Each lab compiles $g_i$ and trains it at screening
fidelity, returning measured relative $L_2$ error on clean and noisy data.
\item \textit{Peer review.} Labs review a sample of peers and cast soft votes;
votes accumulate as citation-like influence and update trust scores.
\item \textit{Lifecycle.} A composite fitness score combines accuracy, generalization,
parameter efficiency, and novelty (with an accuracy floor); the global best is
updated, influential labs persist and breed, and uncited labs are 
replaced.
\end{enumerate}

The composite fitness score for virtual lab $i$ is
\begin{equation}
F_i \,=\, w_a\,A_i + w_g\,G_i + w_e\,E_i + w_n\,\nu_i\,,
\end{equation}
where $(w_a,w_g,w_e,w_n)=(0.70,0.20,0.05,0.05)$. The fitness is zeroed when accuracy falls below a floor, which prevents trivially small or degenerate models from winning on efficiency or novelty alone. Here $A_i$ is an accuracy score derived from the relative $L_2$ error, $G_i$ rewards robustness to input noise, $E_i$ rewards parameter parsimony, and $\nu_i$ is an architectural-novelty score relative to the rest of the population. The weights are fixed a priori, shared by every run and both conditions, and were not tuned; accuracy dominates by design. Making the weights adaptive is left to future work.

\begin{algorithm}[t]
\caption{Neural operator AI scientific community.}
\label{alg:swarm}
\begin{algorithmic}[1]
\State initialize $N$ labs across paradigms (FNO, DeepONet, attention, wavelet, convolutional, hybrid).
\For{iteration $t=1,\dots,T$}
  \For{each active lab $i$} 
     \State $g_i \gets \textsc{Plan}(g_i,\, g^{\text{best}},\, \{g_j\}_{j\neq i},\, \text{history}_i)$
  \EndFor
  \For{each active lab $i$} 
     \State $\theta_i \gets \textsc{Train}(g_i)$; \; measure relative $L_2$
  \EndFor
  \State $\{v_i\} \gets \textsc{PeerReview}(\{(g_i,\text{fitness}_i)\})$
  \State update composite fitness, trust, global best; deactivate  uncited labs, breed influential labs
\EndFor
\end{algorithmic}
\end{algorithm}


\paragraph{Individual virtual labs.}
Each virtual lab consists of three agents. The \textbf{planner} proposes the next genome; the \textbf{worker} compiles and trains the corresponding operator with gradient descent, returning measured relative $L_2$ error; the \textbf{reviewer} casts peer-review votes over a sample of peer candidates. The planner queries an LLM for the next genome, given its own history, the global best, and a summary of peer labs and their measured fitness, while the reviewer queries an LLM to score candidate genomes against measured fitness. Proposals are parsed from JSON and snapped to the valid genome space. We serve \texttt{gemma3:12b} locally via Ollama, so the pipeline is reproducible with no paid API. The worker is deliberately numerical and not an LLM, so that the fitness signal is consistent and comparable across labs and iterations. 

\paragraph{LLM ablation study.}
The framework is model-agnostic in the planner and reviewer roles, which lets us run a clean ablation study of LLM agency, by replacing the LLM agents by rule-based alternatives, where the planner uses explore/exploit particle-swarm updates, while the reviewer uses accuracy-proportional voting. No language model is involved. The rule-based alternatives share the same swarm size, iteration count, training budget, fitness, and random seeds as the original AI scientific community; only the planner/reviewer policy differs. This isolates the effect of LLM agency.

\paragraph{Agent auditing framework.}\label{sec:evidence}
A recurring risk in agentic systems is that a language model silently degrades to a rule-based fallback (e.g.\ on a transient error or malformed output). We guard against this by logging every LLM call, namely, system prompt, user prompt, raw response, role, and parse status, and by auditing each run's log for planner/reviewer fallbacks. Table~\ref{tab:evidence} reports, per LLM run, the planner and reviewer call counts, the parse-success rate, and the fallback count. A fallback count of zero certifies that planning and review were LLM-driven throughout; the transcripts are released so this audit is independently reproducible. Auditing for this silent-degradation failure mode is not yet common practice in agentic pipelines, so we report the fallback count for verifiability rather than as a headline result. \footnote{At first, we adopted \texttt{gemma4:e4b}, a smaller efficiency-tier LLM, which ran into the silent non-LLM fallback issue, even with prompt hardening and an enlarged context window. This led us to switch to \texttt{gemma3:12b} and audit the LLM calls.}

\paragraph{Screening fidelity.}
Each candidate is trained at deliberately reduced ``screening'' fidelity (small data, few epochs, modest resolution) so a single operator trains in seconds, not hours, making a swarm of hundreds of trainings tractable on one GPU (Section~\ref{sec:cost}). Absolute screening errors are therefore higher than fully-converged training.

\section{Results}\label{sec:results}

\paragraph{Experimental setup.}\label{sec:setup}
We consider five problems: piecewise regression, the linear advection and Burgers 1D PDEs, and the Navier-Stokes (in vorticity form with Kolmogorov forcing) and Darcy flow 2D PDEs. Each simulation is repeated three times, correspoding to pytorch random number generator seeds ($42,137,2024$) in both the LLM and the rule-based ablated configuration ($5\times2\times3=30$ runs), with $N=16$ labs for $T=20$ iterations. All runs are resume-safe (checkpointed per iteration) to survive interruptions. We additionally train baselines to convergence (DeepONet, POD-DeepONet, FNO, and Transformer) with input/output normalization in each simulation. 
All reported timings are on a single NVIDIA GPU RTX~4080. See Appendix~\ref{app:details} for further details.

\paragraph{What the LLM community discovers.}
Every LLM winner is a multi-family hybrid, and the mixture is regime-appropriate; e.g., they tend to lead the stacks with Fourier blocks in the spectral-friendly advection, Burgers and Navier-Stokes problems, as can be seen in Figure~\ref{fig:fig_archdiagram.pdf}   (see Table~\label{app:perseed} in Appendix for the complete list of winning architectures). Table~\ref{tab:baselines} reports the accuracy of the two winning architectures for the two most challenging problems, namely the 2D NS and Darcy PDEs, each retrained from scratch and evaluated on both problems, compared to the baselines. 
The Darcy hybrid attains $0.0631\pm0.0083$ relative $L_2$ error, the best mean Darcy error (averaged over the three simulations), statistically tied with DeepONet ($0.0649\pm0.0037$). The NS hybrid reaches $0.0005$ relative $L_2$ error, well below every non-FNO baseline and on par with FNO ($0.0002$), both reaching near-perfect accuracy on this periodic, spectral-friendly problem. For the trade-off between accuracy and parameter count, see Figure~\ref{fig:fig_pareto.pdf} in Appendix~\ref{app:pareto}.

\figfile{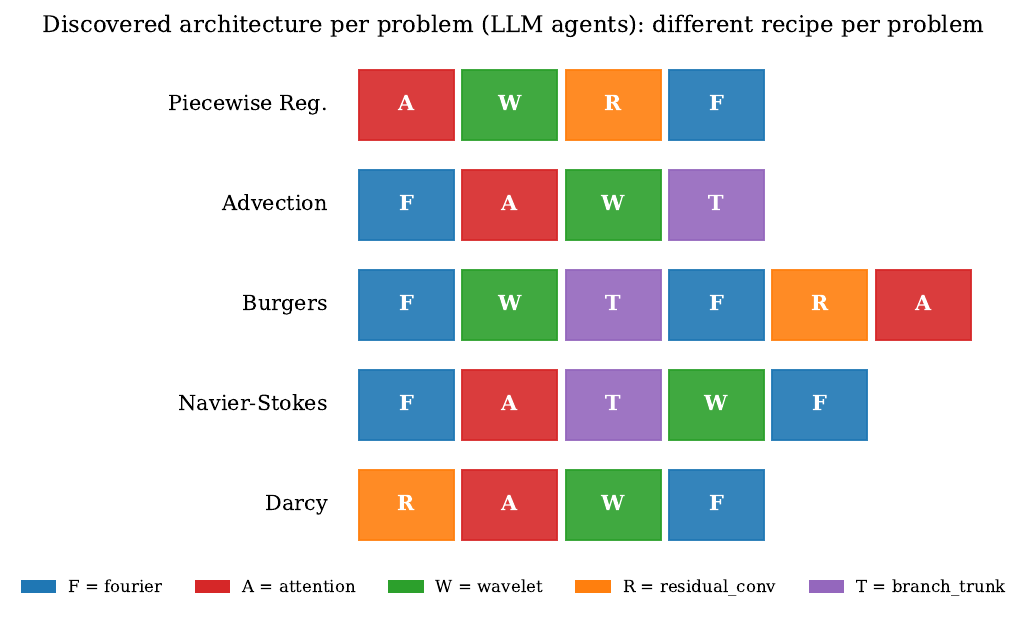}{0.85}{Winning architectures in one of the simulations (random seed 42). Every winner is a multi-family hybrid. As expected, Fourier blocks lead the stacks in the spectral-friendly advection, Burgers and Navier-Stokes problems.}{fig:fig_archdiagram.pdf}

\paragraph{How the LLM plans.}
The transcripts let us inspect the planner's \emph{decisions}, not only its outcomes. Across 4{,}817 planner calls the LLM chose \textsc{hybridize} in 99.8\% of decisions (the remainder \textsc{explore}; never \textsc{exploit}), and its rationales condition on the summarized swarm state, e.g.\ ``\emph{several labs are converging on FNO and transformer architectures\ldots\ I'll introduce a hybrid approach by incorporating attention blocks}'' (Navier-Stokes problem, random seed 42). This action bias is the mechanism behind the hybrids above: a planner that almost always grafts blocks from other families structurally produces multi-family hybrids. The bias is partly prompt-induced (the planner is instructed to preserve diversity when the swarm converges; see Appendix~\ref{app:prompts}). 

\paragraph{What the reviewer rewards.}
The same transcripts quantify the reviewer. Across 4{,}804 scoreable review calls, the rank correlation between the reviewer's vote distribution and the candidates' measured accuracy is $\bar{\rho}=0.64$ (Spearman; positive in 88\% of calls, a perfect accuracy ordering in 50\%), so peer review is grounded in the fitness signal. Its rationales cite accuracy in 97\% of calls, parameter efficiency in 83\%, and robustness in 71\%, but novelty in only 26\%: the deviations from pure accuracy ordering are consistent with the prompt's instruction to also reward novelty and efficiency. Review is thus fitness-anchored with a diversity tilt.



\paragraph{No universal winner.}
We can see in Table~\ref{tab:baselines} that no architecture dominates uniformly over both problems. In addition, the discovered Navier-Stokes  hybrid does badly in the Darcy problem, while the discovered Darcy hybrid is about $3.7\times$ worse than the NS hybrid on the NS problem. The results confirm what is already known in the literature, that FNO dominates the periodic, spectral-friendly Navier-Stokes problem, while DeepONet-style models beat FNO on the elliptic Darcy problem. 

\begin{table}[t]\centering
\caption{Relative $L_2$ error for the winning architectures, each retrained from scratch, and the baselines, for the two most challenging problems, namely the 2D NS and Darcy PDEs. Mean and standard deviations over the three simulations are reported. The Darcy hybrid attains the best mean accuracy, statistically tied with DeepONet, while the NS hybrid is on par with FNO, at a smaller number of parameters, and well below every non-FNO baseline.}
\label{tab:baselines}
\inputif{tab_baselines.tex}{(baselines table is generated by the build pipeline.)}
\end{table}

\paragraph{Ablation study: what does LLM agency add?}
Table~\ref{tab:discovery} reports the discovered error and parameter count per problem under LLM and rule-based coordination, while Figure~\ref{fig:fig_ablation.pdf} compares discovered relative $L_2$ errors, and Figure~\ref{fig:fig_convergence.pdf} compares convergence of the swarm's best errors per iteration. Because both conditions train identical operators under an identical budget, any difference is attributable to the planner/reviewer policy. The screening accuracies are statistically indistinguishable on all five problems (overlapping within $1$ s.d.) over the three simulations. We do not claim an accuracy win for LLM agency. The measurable difference is architectural: the LLM discovers parameter-lean multi-family hybrids where the rule-based condition instead tends to collapse to canonical, nearly single-family designs; e.g. pure-Fourier stacks on Navier-Stokes (the textbook answer for a periodic, spectrally friendly PDE) and wavelet-dominated stacks on Darcy and piecewise regression. On Navier-Stokes the LLM hybrids match the pure-spectral rule-based accuracy with $2.5\times$ fewer parameters on average ($2.2$M vs.\ $5.6$M), and it does so reproducibly (the identical advection recipe across all simulations). 

\begin{table}[t]\centering
\caption{Ablation study results. Relative $L_2$ error on the clean data and parameter
count, LLM vs.\ rule-based, mean$\pm$std over the three simulations.}
\label{tab:discovery}
\inputif{tab_discovery.tex}{(discovery table populates as the campaign completes.)}
\end{table}

\figfile{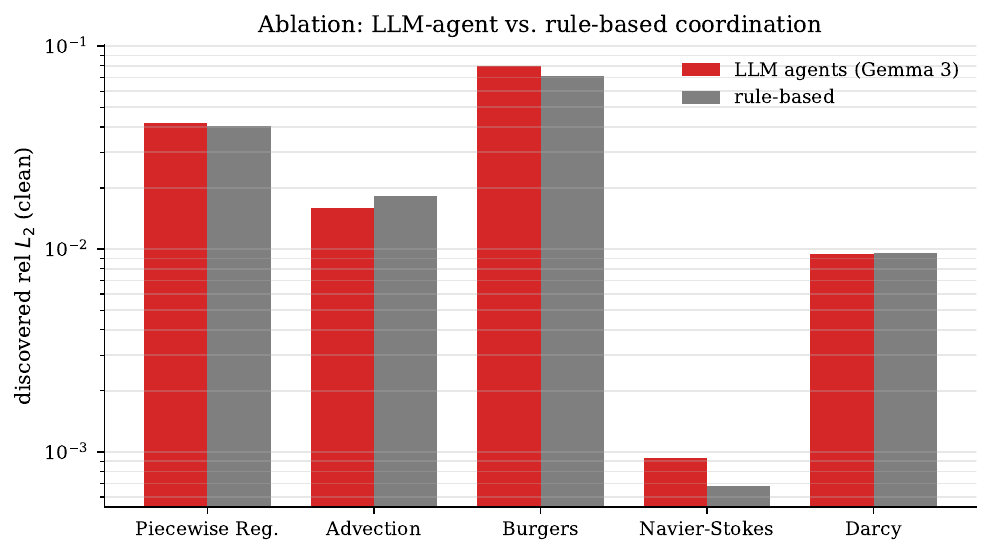}{0.7}{Ablation study results. Discovered relative $L_2$ error per problem: LLM vs.\ rule-based, under an identical training budget.}{fig:fig_ablation.pdf}

\figfile{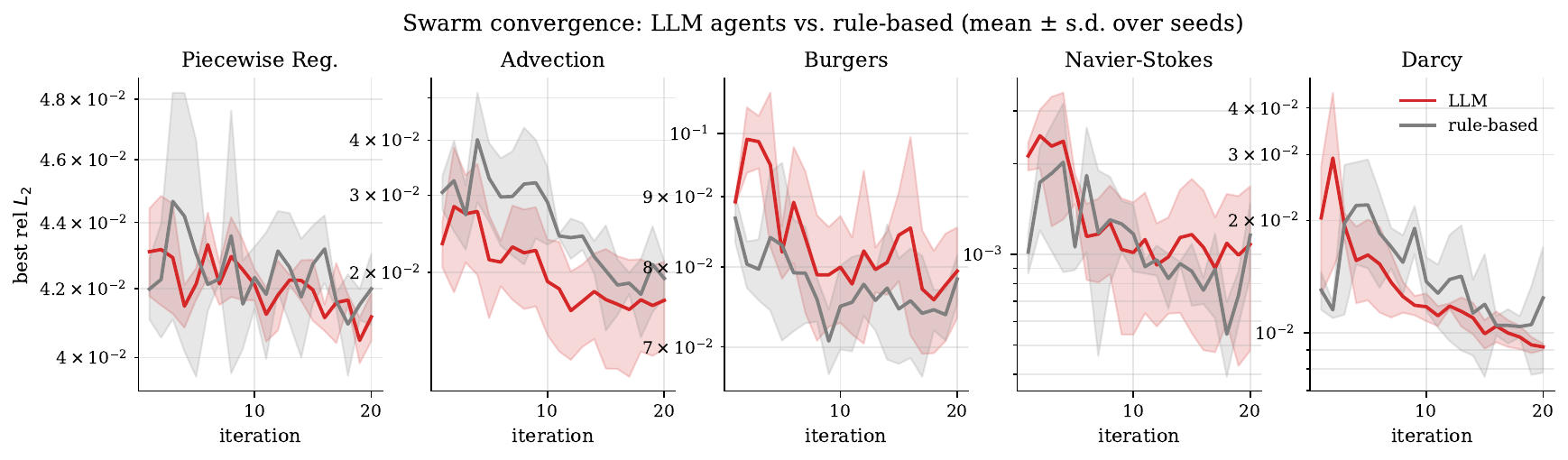}{1.0}{Ablation study results. Best relative $L_2$ error per iteration, LLM vs.\ rule-based, mean $\pm$ s.d.\ over simulations.}{fig:fig_convergence.pdf}

\paragraph{Agent auditing results.}
Table~\ref{tab:evidence} summarizes the transcript audit over all LLM runs: planner and reviewer calls, parse-success, and fallbacks to rule-based. Across the campaign we observe zero rule-based fallbacks, certifying that planning and review were LLM-driven.

\begin{table}[t]\centering
\caption{Agent Audit. Planner/reviewer LLM call counts per run, parse-success, and fallbacks. Zero fallbacks certifies LLM-driven planning and review. Counts above the nominal $16\times20=320$ reflect re-planned iterations after mid-run interruptions (the runs are checkpointed and resume-safe); every call, including repeats, is in the released transcripts.}
\label{tab:evidence}
\inputif{tab_evidence.tex}{(evidence table populates as the campaign completes.)}
\end{table}

\section{Computational Cost}\label{sec:cost}
Table~\ref{tab:timing} and Figure~\ref{fig:fig_timing.pdf} report wall-clock times for a full simulation run and the time per lab evaluation, for LLM and rule-based labs. Because the LLM and rule-based AI scientific communities train identical operators under an identical budget, the difference in per-lab time is precisely the overhead of LLM-driven planning and review (two language-model calls per lab per iteration): roughly $14$ sec per lab evaluation in the 1D problems and $25$-$33$s in the 2D problems, i.e.\ a $2$-$4\times$ increase in total wall-clock time over rule-based coordination. The LLM labs also carry higher run-to-run variance in the 2D problems (e.g.\ $271\pm104$ min on Navier-Stokes): a consequence of letting the planner commit to parameter-heavy hybrids whose training dominates the budget. 

\begin{table}[t]\centering
\caption{Computational cost on a single RTX~4080: wall-clock simulation and per lab evaluation time, LLM vs.\ rule-based, mean$\pm$std over simulations. The LLM-rule-based
gap reveals the cost of using LLMs.}
\label{tab:timing}
\inputif{tab_timing.tex}{(timing table populates as the campaign completes.)}
\end{table}

\figfile{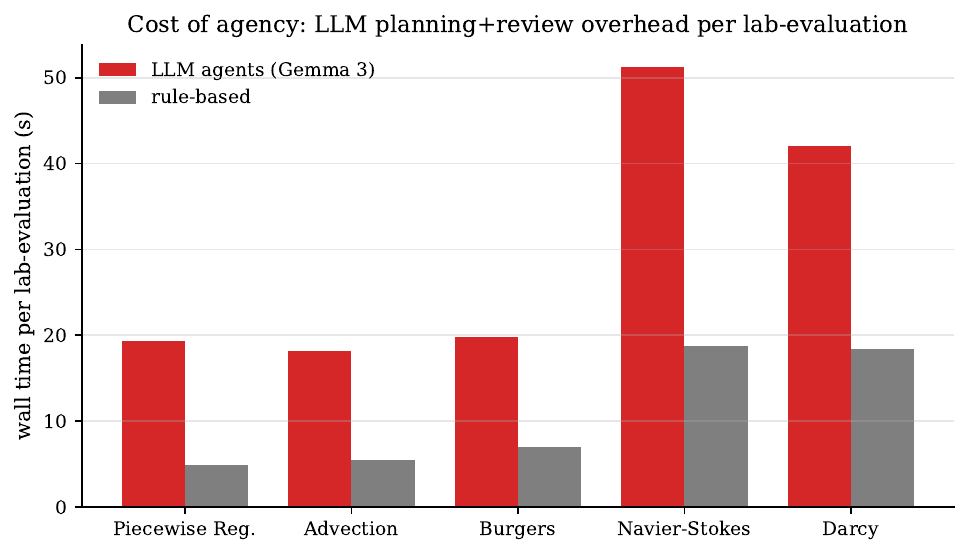}{0.7}{Per lab-evaluation wall time, LLM vs.\ rule-based. The gap
is the cost of LLM planning and peer review at identical training budget.}{fig:fig_timing.pdf}

\paragraph{Why a swarm of hundreds of trainings is feasible.}
The swarm screens architectures at reduced fidelity, so a single operator trains in seconds, not hours. Table~\ref{tab:trainop} lists single-operator training times to convergence at screening fidelity in the Navier-Stokes problem; cost is dominated by epochs and block type, not parameter count (the 105K-parameter transformer is slower to train than the 1.2M-parameter FNO because attention is quadratic in sequence length). A full simulation run evaluates $16\times20=320$ architectures within the budgets above, reserving expensive full-fidelity training for the final discovered architecture and the baseline comparison. This follows the standard multi-fidelity procedure to screen cheaply and commit expensively.

\begin{table}[t]\centering
\caption{Single-operator training time at screening fidelity in the Navier-Stokes problem.}
\label{tab:trainop}
\inputif{tab_trainop.tex}{(training-time table is generated by the build pipeline.)}
\end{table}

\pagebreak
\section{Limitations}
\paragraph{LLM-as-policy, not autonomous tool use.} Our agents use the LLM as a
\emph{policy} for the planning and review decisions; the LLM does not execute code or call external tools. Including fully-agentic virtual labs as in~\cite{lu2026aiscientist,toscano2025athena} is left to future work.

\paragraph{Literature priors.} A mid-size or frontier LLM has read the
operator-learning literature and may ``know'', for example, that FNO is suitable for periodic, spectral-friendly transport problems. This is simultaneously a feature (informed priors) and a confounding factor (it is not prior-free discovery); the rule-based alternative in the ablation study provides the prior-free reference point.
\paragraph{Model scale.} We use a single local model (\texttt{gemma3:12b}).
Whether a frontier model changes discovery quality is left to future work.
\paragraph{LLM planning is stochastic.} We report results for three independent random simulation runs and
release transcripts rather than claiming bit-for-bit reproducibility.
\paragraph{Screening fidelity.} The swarm screens at reduced fidelity for computational cost reasons.
\paragraph{Joint role ablation.} The ablation study switches planner and reviewer
together. Disentangling their individual contributions requires a factorial design, which we leave to future work.

\section{Reproducibility}\label{sec:repro}
The resume-safe runner, exact configurations, and the complete per-run LLM transcripts are available at \url{https://github.com/luislootx/AI-SC}. Every reported LLM run has its \texttt{llm\_transcript.jsonl} file, and the agent audit (Table~\ref{tab:evidence}) is reproducible directly from these files. The LLM backend (\texttt{gemma3:12b} via Ollama) runs locally with no paid API. Representative planner and reviewer prompts are listed in Appendix~\ref{app:prompts}.

\section{Conclusion}
We presented an agentic LLM-driven neural operator AI scientific community. Each virtual lab in the community consists of a planner, a worker, and a reviewer. The planner and reviewer are LLMs, while the worker is numerical (gradient-based), by design, so the fitness signal stays comparable across the swarm. All 9{,}623 LLM calls are logged and audited, verifying that no fallbacks to rule-based alternatives occur. We show that this AI scientific community is capable of discovering high-accuracy, low-parameter neural operator architectures at a quantified LLM overhead. We conducted a controlled ablation study against rule-based coordination, which shows that the LLM agents maintain diversity, while the rule-based agents collapse to pure stacks. Future work will include frontier LLM backends, fully autonomous agents, more roles per virtual lab, a factorial role ablation study, larger block vocabularies (e.g., graph operators, neural Galerkin), action-space debiasing of the planner, and full-fidelity re-training of every discovered architecture.

\begin{ack}
We thank the organizers of the ICERM Hot Topics Workshop on Agentic Scientific Computing and Scientific Machine Learning, where a preliminary version of this work was presented as a poster and valuable feedback was received, and the Texas A\&M ECEN graduate program for travel support.
\end{ack}

\bibliographystyle{unsrt}
\bibliography{refs}

\begin{thebibliography}{10}

\bibitem{kovachki2023neuraloperator}
Nikola Kovachki, Zongyi Li, Burigede Liu, Kamyar Azizzadenesheli, Kaushik
  Bhattacharya, Andrew Stuart, and Anima Anandkumar.
\newblock Neural operator: Learning maps between function spaces with
  applications to {PDEs}.
\newblock {\em Journal of Machine Learning Research}, 24(89):1--97, 2023.

\bibitem{li2021fno}
Zongyi Li, Nikola Kovachki, Kamyar Azizzadenesheli, Burigede Liu, Kaushik
  Bhattacharya, Andrew Stuart, and Anima Anandkumar.
\newblock Fourier neural operator for parametric partial differential
  equations.
\newblock In {\em International Conference on Learning Representations (ICLR)},
  2021.

\bibitem{lu2021deeponet}
Lu~Lu, Pengzhan Jin, Guofei Pang, Zhongqiang Zhang, and George~Em Karniadakis.
\newblock Learning nonlinear operators via {DeepONet} based on the universal
  approximation theorem of operators.
\newblock {\em Nature Machine Intelligence}, 3(3):218--229, 2021.

\bibitem{tripura2023wno}
Tapas Tripura and Souvik Chakraborty.
\newblock Wavelet neural operator for solving parametric partial differential
  equations in computational mechanics problems.
\newblock {\em Computer Methods in Applied Mechanics and Engineering},
  404:115783, 2023.

\bibitem{cao2021galerkin}
Shuhao Cao.
\newblock Choose a transformer: Fourier or galerkin.
\newblock In {\em Advances in Neural Information Processing Systems (NeurIPS)},
  volume~34, 2021.

\bibitem{hao2023gnot}
Zhongkai Hao, Zhengyi Wang, Hang Su, Chengyang Ying, Yinpeng Dong, Songming
  Liu, Ze~Cheng, Jian Song, and Jun Zhu.
\newblock {GNOT}: A general neural operator transformer for operator learning.
\newblock In {\em Proceedings of the 40th International Conference on Machine
  Learning (ICML)}, volume 202, 2023.

\bibitem{braganeto2026aicommunity}
Ulisses Braga-Neto.
\newblock The {AI} scientific community: Agentic virtual lab swarms.
\newblock {\em arXiv preprint arXiv:2603.21344}, 2026.

\bibitem{lulu2022comparison}
Lu~Lu, Xuhui Meng, Shengze Cai, Zhiping Mao, Somdatta Goswami, Zhongqiang
  Zhang, and George~Em Karniadakis.
\newblock A comprehensive and fair comparison of two neural operators (with
  practical extensions) based on {FAIR} data.
\newblock {\em Computer Methods in Applied Mechanics and Engineering},
  393:114778, 2022.

\bibitem{raonic2023convolutional}
Bogdan Raonic, Roberto Molinaro, Tobias Rohner, Siddhartha Mishra, and Emmanuel
  de~Bezenac.
\newblock Convolutional neural operators.
\newblock In {\em ICLR 2023 workshop on physics for machine learning}, 2023.

\bibitem{rahman2023uno}
Md~Ashiqur Rahman, Zachary~E. Ross, and Kamyar Azizzadenesheli.
\newblock {U-NO}: U-shaped neural operators.
\newblock {\em Transactions on Machine Learning Research (TMLR)}, 2023.
\newblock arXiv:2204.11127.

\bibitem{li2023geofno}
Zongyi Li, Daniel~Zhengyu Huang, Burigede Liu, and Anima Anandkumar.
\newblock Fourier neural operator with learned deformations for {PDEs} on
  general geometries.
\newblock {\em Journal of Machine Learning Research}, 24(388):1--26, 2023.
\newblock arXiv:2207.05209.

\bibitem{zoph2017nasrl}
Barret Zoph and Quoc~V. Le.
\newblock Neural architecture search with reinforcement learning.
\newblock In {\em International Conference on Learning Representations (ICLR)},
  2017.

\bibitem{real2019regevo}
Esteban Real, Alok Aggarwal, Yanping Huang, and Quoc~V. Le.
\newblock Regularized evolution for image classifier architecture search.
\newblock In {\em Proceedings of the AAAI Conference on Artificial
  Intelligence}, volume~33, pages 4780--4789, 2019.

\bibitem{liu2019darts}
Hanxiao Liu, Karen Simonyan, and Yiming Yang.
\newblock {DARTS}: Differentiable architecture search.
\newblock In {\em International Conference on Learning Representations (ICLR)},
  2019.
\newblock arXiv:1806.09055.

\bibitem{elsken2019nassurvey}
Thomas Elsken, Jan~Hendrik Metzen, and Frank Hutter.
\newblock Neural architecture search: A survey.
\newblock {\em Journal of Machine Learning Research}, 20(55):1--21, 2019.

\bibitem{real2020automlzero}
Esteban Real, Chen Liang, David~R. So, and Quoc~V. Le.
\newblock {AutoML-Zero}: Evolving machine learning algorithms from scratch.
\newblock In {\em Proceedings of the 37th International Conference on Machine
  Learning (ICML)}, volume 119 of {\em Proceedings of Machine Learning
  Research}, pages 8007--8019, 2020.
\newblock arXiv:2003.03384.

\bibitem{kennedy1995pso}
James Kennedy and Russell Eberhart.
\newblock Particle swarm optimization.
\newblock In {\em Proceedings of the IEEE International Conference on Neural
  Networks (ICNN'95)}, volume~4, pages 1942--1948, 1995.

\bibitem{yang2024opro}
Chengrun Yang, Xuezhi Wang, Yifeng Lu, Hanxiao Liu, Quoc~V. Le, Denny Zhou, and
  Xinyun Chen.
\newblock Large language models as optimizers.
\newblock In {\em International Conference on Learning Representations (ICLR)},
  2024.

\bibitem{zheng2023genius}
Mingkai Zheng, Xiu Su, Shan You, Fei Wang, Chen Qian, Chang Xu, and Samuel
  Albanie.
\newblock Can {GPT-4} perform neural architecture search?
\newblock {\em arXiv preprint arXiv:2304.10970}, 2023.

\bibitem{romeraparedes2024funsearch}
Bernardino Romera-Paredes, Mohammadamin Barekatain, Alexander Novikov, Matej
  Balog, M.~Pawan Kumar, Emilien Dupont, Francisco J.~R. Ruiz, Jordan~S.
  Ellenberg, Pengming Wang, Omar Fawzi, Pushmeet Kohli, and Alhussein Fawzi.
\newblock Mathematical discoveries from program search with large language
  models.
\newblock {\em Nature}, 625(7995):468--475, 2024.

\bibitem{chen2023evoprompting}
Angelica Chen, David Dohan, and David So.
\newblock {EvoPrompting}: Language models for code-level neural architecture
  search.
\newblock In {\em Advances in Neural Information Processing Systems 36
  (NeurIPS)}, 2023.
\newblock arXiv:2302.14838.

\bibitem{wu2024autogen}
Qingyun Wu, Gagan Bansal, Jieyu Zhang, Yiran Wu, Shaokun Zhang, Erkang Zhu,
  Beibin Li, Li~Jiang, Xiaoyun Zhang, and Chi Wang.
\newblock {AutoGen}: Enabling next-gen {LLM} applications via multi-agent
  conversation.
\newblock In {\em First Conference on Language Modeling (COLM)}, 2024.
\newblock arXiv:2308.08155.

\bibitem{li2023camel}
Guohao Li, Hasan Abed Al~Kader Hammoud, Hani Itani, Dmitrii Khizbullin, and
  Bernard Ghanem.
\newblock {CAMEL}: Communicative agents for ``mind'' exploration of large
  language model society.
\newblock In {\em Advances in Neural Information Processing Systems 36
  (NeurIPS)}, 2023.
\newblock arXiv:2303.17760.

\bibitem{qian2024chatdev}
Chen Qian, Wei Liu, Hongzhang Liu, Nuo Chen, Yufan Dang, Jiahao Li, Cheng Yang,
  Weize Chen, Yusheng Su, Xin Cong, Juyuan Xu, Dahai Li, Zhiyuan Liu, and
  Maosong Sun.
\newblock {ChatDev}: Communicative agents for software development.
\newblock In {\em Proceedings of the 62nd Annual Meeting of the Association for
  Computational Linguistics (ACL)}, 2024.
\newblock arXiv:2307.07924.

\bibitem{hong2024metagpt}
Sirui Hong, Mingchen Zhuge, Jonathan Chen, Xiawu Zheng, Yuheng Cheng, Ceyao
  Zhang, Jinlin Wang, Zili Wang, Steven Ka~Shing Yau, Zijuan Lin, Liyang Zhou,
  Chenyu Ran, Lingfeng Xiao, Chenglin Wu, and J{\"u}rgen Schmidhuber.
\newblock {MetaGPT}: Meta programming for a multi-agent collaborative
  framework.
\newblock In {\em International Conference on Learning Representations (ICLR)},
  2024.
\newblock arXiv:2308.00352.

\bibitem{lu2026aiscientist}
Chris Lu, Cong Lu, Robert~Tjarko Lange, Jakob Foerster, Jeff Clune, and David
  Ha.
\newblock Towards end-to-end automation of {AI} research.
\newblock {\em Nature}, 651(8107):914--919, 2026.

\bibitem{toscano2025athena}
Juan~Diego Toscano, Daniel~T. Chen, and George~Em Karniadakis.
\newblock {ATHENA}: Agentic team for hierarchical evolutionary numerical
  algorithms.
\newblock {\em arXiv preprint arXiv:2512.03476}, 2025.

\bibitem{zheng2023llmjudge}
Lianmin Zheng, Wei-Lin Chiang, Ying Sheng, Siyuan Zhuang, Zhanghao Wu, Yonghao
  Zhuang, Zi~Lin, Zhuohan Li, Dacheng Li, Eric~P. Xing, Hao Zhang, Joseph~E.
  Gonzalez, and Ion Stoica.
\newblock Judging {LLM}-as-a-judge with {MT-Bench} and chatbot arena.
\newblock In {\em Advances in Neural Information Processing Systems 36
  (NeurIPS), Datasets and Benchmarks Track}, 2023.
\newblock arXiv:2306.05685.

\end{thebibliography}

\appendix

\section*{Appendix}

\section{Further Experimental Details}\label{app:details}
Table~\ref{tab:expdetails} lists the screening-phase configuration per problem (from each run in file \texttt{config.json} in the repository). Both LLM and rule-based agents share these budgets exactly; only the planner/reviewer policy differs. Every candidate is trained from scratch for the stated epochs each iteration; robustness to noise is measured by adding Gaussian noise $\varepsilon\sim\mathcal{N}(0,\,0.1^2)$ to the data. The LLM backend is \texttt{gemma3:12b} served by Ollama with an 8192-token context window; all timings are on a single RTX~4080. The separately trained baselines and winning architectures run for 50 epochs for FNO, 80 epochs for the winning hybrids, POD-DeepONet and transformer, and 600 epochs for the DeepONet with input/output normalization, as encoded in file  \texttt{validate\_baselines\_v3.py} in the repository.

\begin{table}[h!]\centering
\caption{Screening-phase configuration per problem (identical across conditions and
simulations).}
\label{tab:expdetails}
\begin{tabular}{lccccc}
\toprule
Problem & Dim & Resolution & Train / test & Epochs per candidate \\
\midrule
Piecewise Reg. & 1D & 128 & 256 / 64 & 40 \\
Advection & 1D & 128 & 256 / 64 & 40 \\
Burgers & 1D & 128 & 256 / 64 & 40 \\
Navier-Stokes & 2D & $32\times32$ & 512 / 128 & 50 \\
Darcy & 2D & $32\times32$ & 512 / 128 & 50 \\
\bottomrule
\end{tabular}
\end{table}

\section{Complete List of Winning Architectures}\label{app:winning}
Table~\ref{tab:perseed} lists the details for the winning architectures in each simulation (5 problems $\times$ 2 conditions $\times$ 3 random seeds), including the rule-based agents in the ablation study. We observe that the LLM agents return diverse architectures, but are problem-specific; e.g. they almost always lead the stacks with Fourier blocks in the spectral-friendly advection, Burgers, and Navier-Stokes problems. The rule-based agents tend to be less diverse and prefer Fourier blocks in the spectral-friendly problems and wavelet blocks in the piecewise regression and Darcy problems. A machine-readable version of these results is in file \texttt{results/aggregate\_v2.json} in the repository. 

\begin{table}[h]\centering
\caption{Winning architectures. Blocks abbreviated as
T=DeepONet, F=Fourier, A=Attention, W=Wavelet, R=Residual Convolution.}
\label{tab:perseed}
{\small
\inputif{tab_perseed.tex}{(per-seed table is generated by the build pipeline.)}
}
\end{table}

\section{Accuracy-Parameter Trade-Off}\label{app:pareto}
Figure~\ref{fig:fig_pareto.pdf} displays the trade-off between accuracy and parameter account obtained in the simulation for the more challenging Navier-Stokes and Darcy 2D problems. All 48 labs (16 labs $\times$ 3 simulations) produced multi-family hybrids, consistent with the planner hybridizing in 99.8\% of its decisions. We observe that the discovered hybrids populate a competitive region of the accuracy-parameter plane.

\figfileh{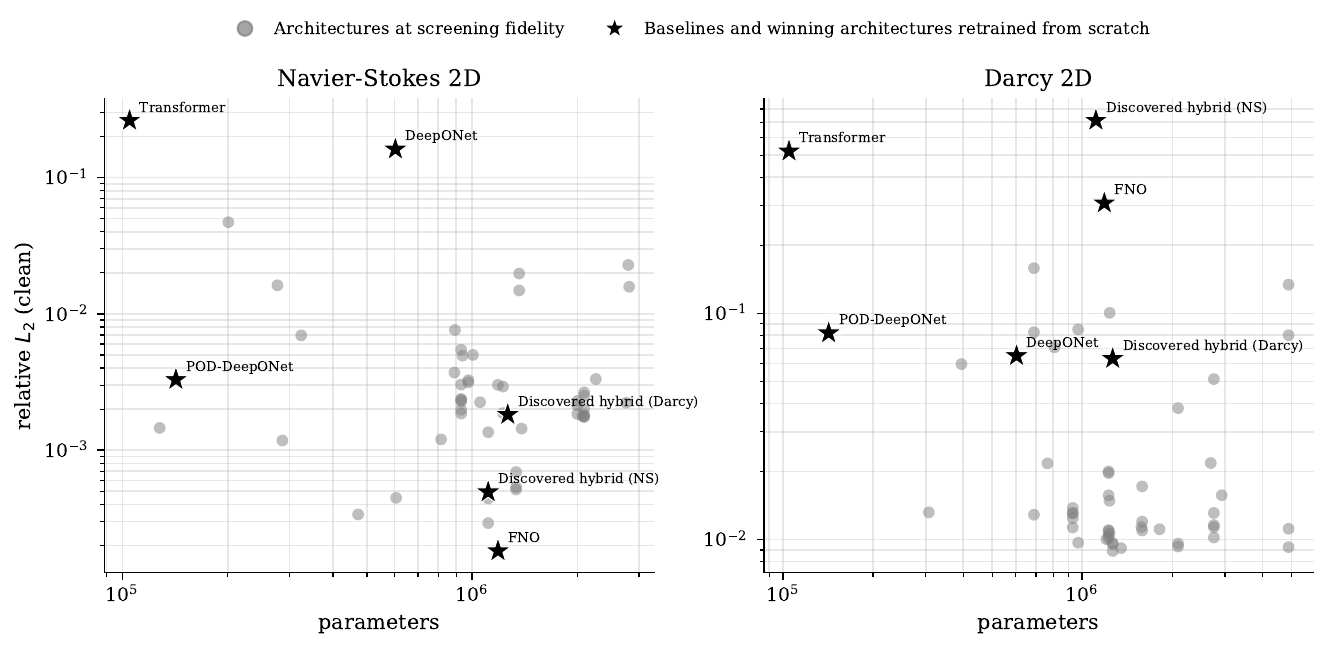}{1.0}{Accuracy-parameter trade-off. problems. The gray dots represent the individual labs, while stars depict the baselines and the winning discovered architectures retrained from scratch to convergence. Error are therefore not directly comparable across the two marker types.}{fig:fig_pareto.pdf}

\section{Agent Prompts}\label{app:prompts}
The planner and reviewer prompts are reproduced verbatim below (the specific problem name is instantiated per run). Note that the planner prompt specifies the genome record to be used by each virtual lab. The outputs are in file \texttt{llm\_transcript.jsonl} in the repository.

\paragraph{Planner prompt.}
{\small\begin{verbatim}
You are the planning agent of a virtual research lab in a
swarm of labs collectively discovering neural operator architectures for 2D
Navier-Stokes. Each lab is one PSO particle; your job is to propose the next
architecture (genome) for your lab to train and evaluate next iteration.

You must balance EXPLORATION (novel hybrids, unexplored block combinations)
with EXPLOITATION (move toward the global best architecture observed in the
swarm). Use the lab's current exploration_rate as a guide: high -> favor
exploration, low -> favor exploitation. Consider what other labs are doing -
if the swarm is converging, propose something different to maintain diversity
and avoid groupthink.

Output STRICT JSON ONLY (no prose, no fences) matching this schema:
{
  "action": "explore" | "exploit" | "hybridize",
  "rationale": "<1-2 sentences explaining your choice>",
  "new_genome": { "block_sequence": [...], "hidden_channels": <int>,
    "fourier_modes": <int>, "activation": "<str>", "use_gating": <bool>,
    "use_skip_connections": <bool>, "dropout_rate": <float>,
    "learning_rate": <float>, "weight_decay": <float> }
}
- ALLOWED_BLOCKS = ["fourier","attention","branch_trunk","wavelet",
                    "residual_conv"]
\end{verbatim}}

\paragraph{Reviewer prompt.}
{\small\begin{verbatim}
You are an anonymous peer reviewer evaluating neural
operator architectures proposed by other virtual labs in a research swarm.

You must rank the candidates on overall scientific merit for solving 2D
Navier-Stokes operator learning. Reward: high accuracy, good generalization
under noise, parameter efficiency, architectural novelty (unusual block
combinations), and theoretical soundness (e.g. fourier blocks for periodic
domains, attention for nonlocal coupling). Penalize: groupthink (identical to
global best), pathological setups (too few or too many blocks, mismatched
modes), or poor measured fitness.

Output STRICT JSON ONLY:
{
  "scores":   { "<lab_id>": <float in [0,1]>, ... },
  "rationale": "<1-2 sentences total, not per lab>"
}
Scores must sum to roughly 1.0 across the candidates (soft-vote distribution).
\end{verbatim}}

\end{document}